%% file: egpaper_final.tex
\documentclass[10pt,twocolumn,letterpaper]{article}

\usepackage{iccv}
\usepackage{times}
\usepackage{epsfig}
\usepackage{graphicx}
\usepackage{amsmath}
\usepackage{amssymb}
\usepackage{times}
\usepackage{epsfig}
\usepackage{graphicx}
\usepackage{amsmath}
\usepackage{amssymb}
\usepackage{bbm}
\usepackage{float}
\usepackage{titlepic}
\usepackage{cuted}
\usepackage{capt-of}
\usepackage{times}
\usepackage{multirow}
\usepackage{colortbl}

\usepackage[pagebackref=true,breaklinks=true,letterpaper=true,colorlinks,bookmarks=false]{hyperref}

\iccvfinalcopy 

\def\iccvPaperID{2322} 

\ificcvfinal\fi

\begin{document}

\title{Person-in-WiFi: Fine-grained Person Perception using WiFi}

\author{Fei Wang\textsuperscript{1,2}\thanks{Work done when at CMU.}~~~~ Sanping Zhou\textsuperscript{1,2}~~~~ Stanislav Panev\textsuperscript{2}~~~~Jinsong Han\textsuperscript{3}~~~~ Dong Huang\textsuperscript{2} \\
\small{\textsuperscript{1}Xi'an Jiaotong University~~~~
\textsuperscript{2}Carnegie Mellon University~~~~
\textsuperscript{3}Zhejiang University~~~~}
\\
\tt\small {feiwang@cmu.edu, sanpingzhou@stu.xjtu.edu.cn, spanev@cmu.edu}\\
\tt\small{hanjinsong@zju.edu.cn, donghuang@cmu.edu}
}
\maketitle

\begin{strip}
\centering
\vspace{-25pt}
\includegraphics[width=1.0\linewidth]{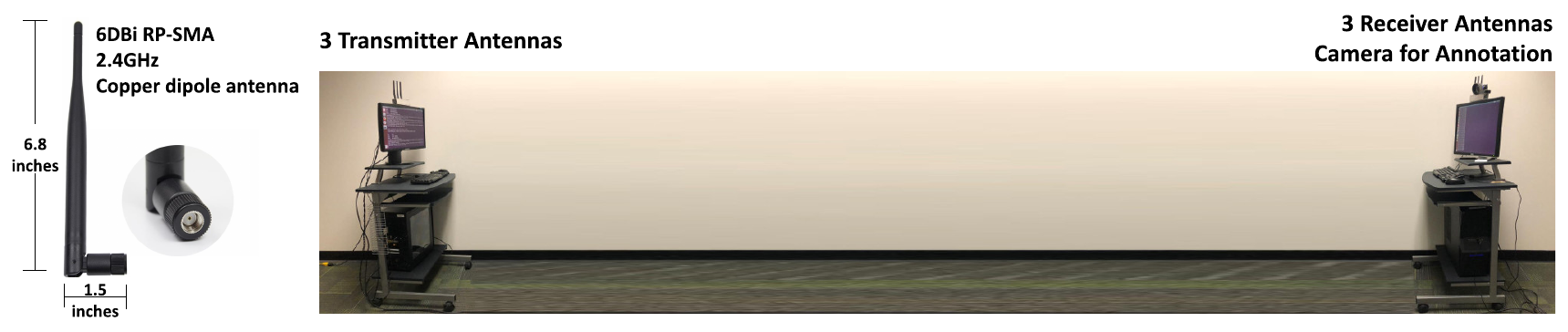}
\includegraphics[width=1.0\linewidth]{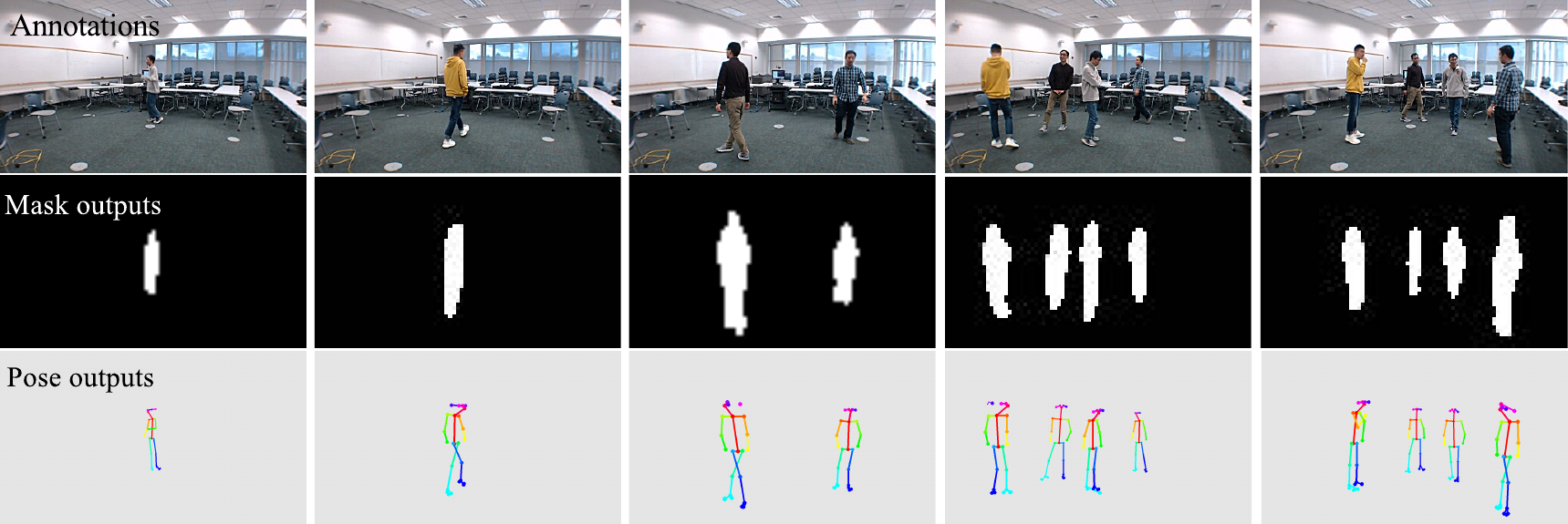}
\captionof{figure}{Person-in-WiFi. Top: WiFi antennas as sensors for person perception. Receiver antennas record WiFi signals as inputs to Person-in-WiFi. The rest rows are, images used to annotate WiFi signals, and two outputs: person segmentation masks and body poses.}\label{fig:first}
\end{strip}

\begin{abstract}


Fine-grained person perception such as body segmentation and pose estimation has been achieved with many 2D and 3D sensors such as RGB/depth cameras, radars (e.g., RF-Pose) and \mbox{LiDARs}. These sensors capture 2D pixels or 3D point clouds of person bodies with high spatial resolution, such that the existing Convolutional Neural Networks can be directly applied for perception. In this paper, we take one step forward to show that fine-grained person perception is possible even with 1D sensors: WiFi antennas. To our knowledge, this is the first work to perceive persons with pervasive WiFi devices, which is cheaper and power efficient than radars and LiDARs, invariant to illumination, and has little privacy concern comparing to cameras. We used two sets of off-the-shelf WiFi antennas to acquire signals, \textit{i.e.}, one transmitter set and one receiver set. Each set contains three antennas lined-up as a regular household WiFi router. The WiFi signal generated by a transmitter antenna, penetrates through and reflects on human bodies, furniture and walls, and then superposes at a receiver antenna as a 1D signal sample (instead of 2D pixels or 3D point clouds).  We developed a deep learning approach that uses annotations on 2D images, takes the received 1D WiFi signals as inputs, and performs body segmentation and pose estimation in an end-to-end manner. Experimental results on over $10^5$ frames under $16$ indoor scenes demonstrate that Person-in-WiFi achieved person perception comparable to approaches using 2D images.
\end{abstract}

\input{tex/introduction.tex}
\input{tex/related.tex}
\input{tex/wifi.tex}
\input{tex/method.tex}
\input{tex/evaluation.tex}

\section{Conclusion}
 WiFi devices as perception sensors are invariant to illumination and privacy-friendly comparing to cameras,  while are cheaper, smaller, and more power efficient than radars and LiDARs. In this paper, we present the first work that given 1D data received at WiFi antennas, it is possible to reconstruct 2D fine-grained spatial information of human bodies. Our Person-in-WiFi approach is based on off-the-shell WiFi antennas lined-up as regular house-hold WiFi routers, making it very easy to develop perception applications in any indoor environments such as warehouse, hospital, office and home.   

\section*{ACKNOWLEDGE}
 We thank Gines Hidalgo and Yaadhav Raaj for their helps in installing and debugging OpenPose Python API. Fei Wang and Sanping Zhou are supported by China Scholarship Council.

{\small
\bibliographystyle{ieee}

\bibliography{egbib}
}

\end{document}

%% file: tex/introduction.tex
\section{Introduction}


To conduct fine-grained person perception like human body segmentation and pose estimation, three main categories of sensors have been used: cameras (2D images), radars (depth maps), and LiDARs (3D point clouds). These approaches require a minimal spatial resolution of sensor outputs. 
For instance, \mbox{$300\times 300$ pixels} images from cameras~\cite{liu2016ssd},
  depth resolution around $2$ cm for radars~\cite{zhao2018through}, or $32$-beam LiDARs~\cite{wang2017characterization,maturana2015voxnet}. Moreover, camera-based solutions are limited by technical challenges such as clothing, background, lighting and occlusion, and social limitations such as privacy concerns. Radar sensors require dedicated hardware, ~\textit{e.g.}, RF-Pose~\cite{zhao2018through} and RF-Capture~\cite{adib2015capturing} used the Frequency Modulated Continuous Wave~(FMCW) technology to produce depth maps, requiring carefully assembled and synchronized $16+4$ T-shaped antenna array with a broad signal bandwidth ($1.78$~GHz). High-definition LiDAR sensors are very expensive and power-consuming, therefore are difficult to apply for daily and household use. 
  


In this paper, we propose a fine-grained person perception solution using WiFi antennas, which is wildly available in warehouse, hospital, office, home where the low illumination, blind spots, privacy issues make cameras not applicable, while radars and LiDARs are too expensive and power-consuming to install. The challenge is that a WiFi antenna can only receive signal as the amplitude of Electromagnetic~(EM) waves. The received amplitude is an one dimensional summary of the 3D space. Reconstructing fine-grained spatial information from the 1D summary is a severely ill-posed problem. It is even more challenging for person perception: (1) WiFi signals are jointly interfered by the human body and environment via the multiple propagation path effect~\cite{yang2015wifi}. (2) Physical differences among different bodies in their bone, muscle and fat distribution~\cite{wang2018csi}. (3) Temporal physical changes such as breath and heartbeats~\cite{wang2016human}. Due to these challenges, WiFi antennas have only been explored preliminarily on detecting the presence or a rough body mass even with a large antenna array \cite{huang2014feasibility, holl2017holography}. To the best of our survey, using WiFi devices on fine-grained person perception has never been addressed. 

To address challenges in this ill-posed problem, our solution generates many 1D samples of the environment and human bodies. Specifically, we used two sets of off-the-shelf WiFi devices, one as transmitter set ($T$) and the other as receiver set ($R$). Three antennas were lined up in each set similar to a standard WiFi router (shown in Figure~\ref{fig:first}). 
WiFi signals were recorded at $30$ frequencies centered at $2.4$~GHz (IEEE~802.11n WiFi communication standard). We recorded RGB videos and computed body segmentation masks and body joints to annotate the signals. 
This setting
provides $9$ propagating pairs among $T$ and $R$ antennas, $30$ 1D superposing patterns per antenna pairs, and multiple 2D spatial annotations of human bodies. We developed a deep learning approach that uses annotations from RGB videos, WiFi samples as input, and reconstructs 2D body segmentation mask and body joint coordinates. Experiments showed that our approach has a comparable ability of person perception as what computer vision approaches can achieve on 2D images. Figure~\ref{fig:first} shows examples of our Person-in-WiFi approach. To our knowledge, this is the first work that demonstrates:
\begin{enumerate}
\item Fine-grained person perception can be achieved using pervasive WiFi antennas. 
\item To sense the human body in 2D, the physical spatial resolution of sensors can be as low as 1 dimension.
\item Deep learning approach mapping WiFi signals to human body segmentation mask and joint coordinates. 
\end{enumerate}

\begin{figure*}[!t]
\centering
\includegraphics[width=1.0\linewidth]{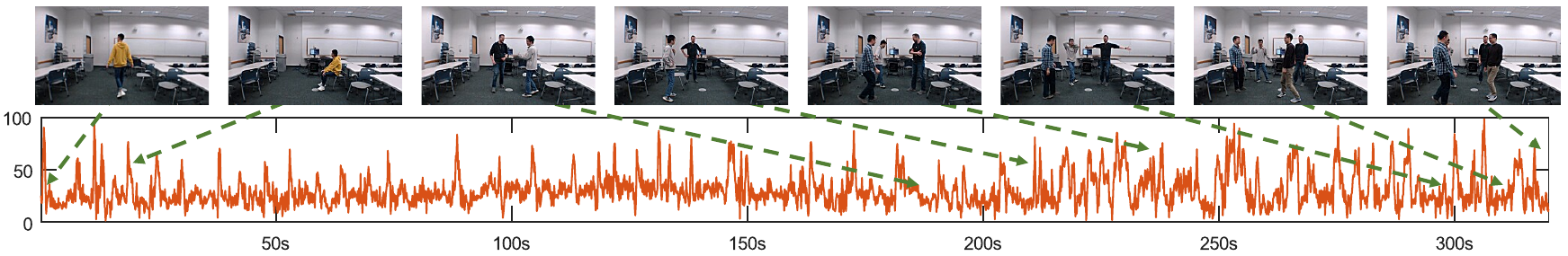}
\caption{WiFi CSI samples recorded during single person moving and multiple person interaction around 320 seconds. The orange curve contains CSI samples of one WiFi signal frequency between one transmitter antenna and one receiver antenna.}\label{fig:csi_series}
\end{figure*} 

%% file: tex/related.tex
\section{Related Work on Person Perception}\label{sec:related}

\textbf{Camera-based.} 
Deep learning has significantly advanced human pose estimation  ~\cite{toshev2014deeppose,tompson2014joint,chen2014articulated,fan2015combining,pishchulin2016deepcut,wei2016convolutional,wei2016convolutional,cao2017realtime} on images captured by monocular cameras, as well as those with optical flow and motion captures \cite{jain2014modeep,fragkiadaki2015recurrent,pfister2015flowing,zhou2016sparseness}. Recent prevalent approaches ~\cite{he2017mask,chen2018cascaded,fang2017rmpe,papandreou2017towards,xiao2018simple} use a powerful person detector such as Faster R-CNN~\cite{ren2015faster}, SSD~\cite{liu2016ssd} Yolo~\cite{redmon2016you}, FPN~\cite{lin2017feature} to crop Region-of-Interest of each person from image feature maps. Then, body-wise pose estimation is done independently on the cropped feature maps. This two-stage schema gains higher performance than previous approaches those are based on global joint heat maps such as OpenPose~\cite{cao2017realtime}.

Unfortunately, we cannot benefit from this two-stage schema because it is not possible to crop 2D pixels of the human body from WiFi signals. Inspired by ~\cite{cao2017realtime}, we developed a deep learning approach to generate Joint Heat Maps~(JHMs) and Part Affinity Fields~(PAFs) directly from WiFi signals. Each JHM encodes one type of joint of all persons, and each PAF encodes the direction and length of person limbs. Then person-wise poses are computed from the JHMs and PAFs similar to~\cite{cao2017realtime}.



\textbf{Radar-based.} Adib et.al.~\cite{adib20143d} introduced a Frequency Modulated Continuous Wave~(FMCW) radar system with broad bandwidth from $5.56$~GHz to $7.25$~GHz for indoor human localization, obtaining a locating resolution of $8.8$~cm. This system is built with the Software-Defined Radar~(SDR) toolkit and T-shaped antenna arrays. Besides, this system is well-synchronized to enable computation on Time-of-Flight~(ToF) of EM wave undergoing transmission, refraction, and reflection, before being received. The ToFs are then used to generate depth maps of the environment. In ~\cite{adib2015capturing}, they promoted the system by focusing on moving person, and generate a rough single person outline with sequential depth maps. Recently, they applied deep learning approaches to do fine-grained human pose estimation using a similar system, termed~RF-Pose~\cite{zhao2018through}. 

\textbf{LiDAR-based.} LiDAR captures 3D point clouds and has been widely used in autonomous robots for Simultaneous Localization and Mapping~(SLAM)~\cite{hess2016real, droeschel2018efficientCS}, person detection~\cite{wang2017characterization,maturana2015voxnet}, tracking~\cite{shaker2008fuzzy,leigh2015person} and surveillance~\cite{benedek20143d,benedek2018lidar,shackleton2010tracking}.
LiDAR sensors provide less spatial resolution than cameras. For instance, a Full HD camera with $90^\circ$ diagonal field-of-view provides an angular resolution of $\approx 0.03^\circ$, whereas the most advanced LiDARs on the market can provide up to $\approx 0.08^\circ$ resolution\footnote{The lower angular resolution the higher spatial resolution.}. Affordable LiDARs usually have at least one magnitude lower angular resolution than the much more affordable cameras. Moreover, LiDARs have sampling rate in the range of $5$-$20$~Hz, which is much lower than other sensors such as cameras ($20$-$60$~Hz) or WiFi adapters ($100$~Hz).
To increase robustness, many researchers combine LiDAR with RGB cameras \cite{premebida2014pedestrian,matti2017combining,han2015realtime} or with motion sensors \cite{costea2017fast} for pedestrian detection.

\textbf{WiFi-based.} To the best of our knowledge, WiFi has been only explored for coarse-grained perception such as 
indoor localization with EM propagating models~\cite{adib2013see,kotaru2015spotfi} and classifying a closed-set of activities, such as opening a door~\cite{pu2013whole}, keystroke~\cite{ali2015keystroke,li2016csi} and dancing~\cite{qian2017inferring}. Wision~\cite{huang2014feasibility} generated a bubble-like 2D heatmap to locate single static person using a $8\times 8$ WiFi antenna array. \cite{holl2017holography} generated the hologram of static objects by sweeping a WiFi antenna in 2D space and recording signals, which virtually simulates a 2D antenna array. 

Till now, fine-grained person perception with WiFi signal, such as body segmentation and pose estimation, has not been well-explored. In this paper, we take one step forward to make this happen.


%% file: tex/wifi.tex
\section{Person Perception with WiFi Signals}

\subsection{Methodology}

\begin{figure}[!t]
\centering
\includegraphics[width=0.95\linewidth]{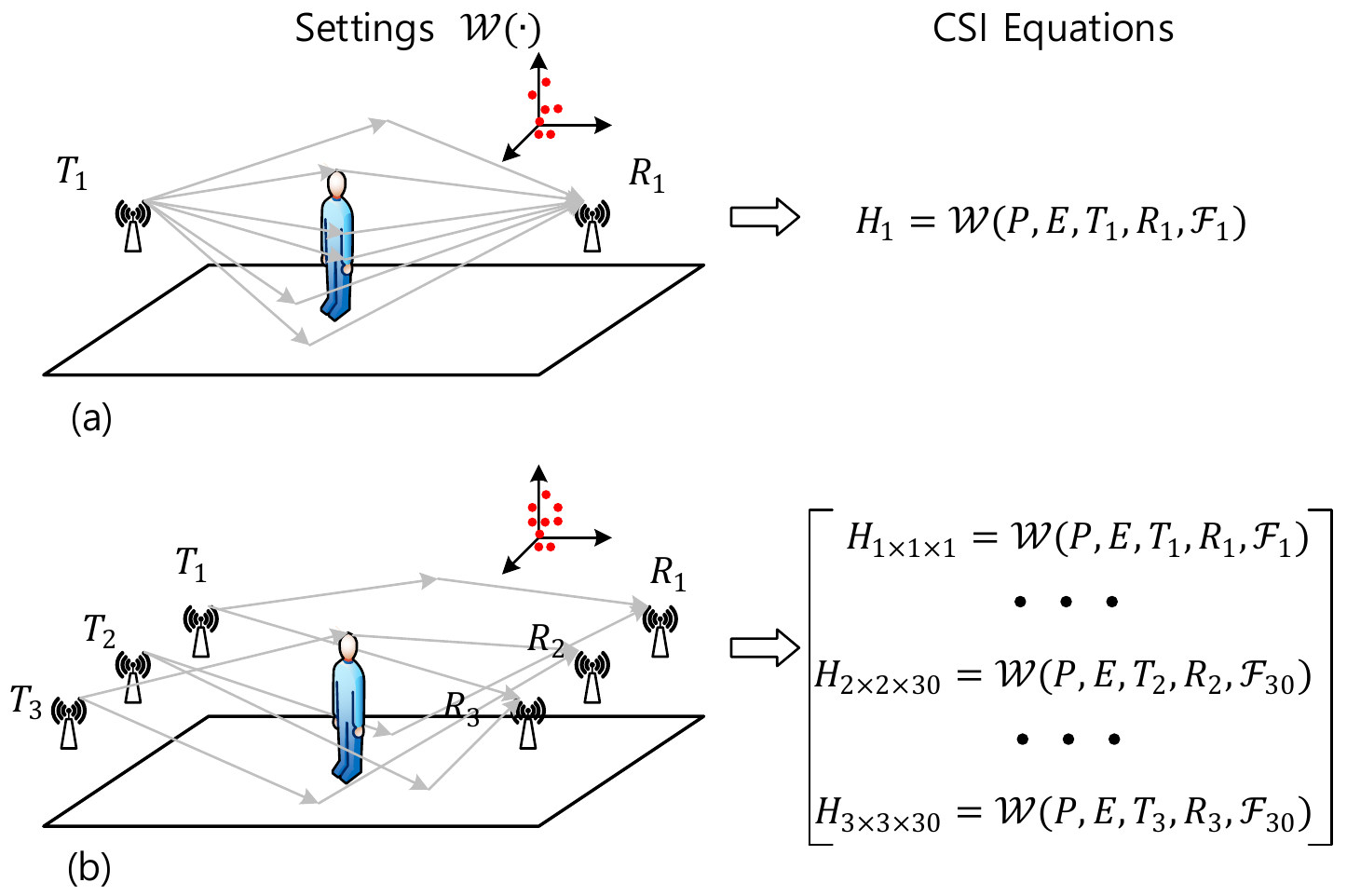}
\caption{WiFi sensing system. $H$: CSI sample, $P$: person body, $E$: environment, $T$: transmitter antenna, $R$: receiver antenna, $F$: EM frequency.}\label{fig:wifi}
\end{figure}

We first consider the simplest setting $\mathcal{W}(\cdot)$ of a WiFi sensing system (Figure~\ref{fig:wifi}~(a)): one transmitting antenna, one receiving antenna and one EM frequency. A person stands still between two antennas and one pulse signal is broadcast from the transmitting antenna. Due to the different EM properties of the human body from the floor, ceiling, furniture and walls, the signal penetrates, refracts and reflects at countless points and directions on the body. This process may probe rich spatial information of both human body ($P$) and environment ($E$) for person perception.


Unfortunately, when the penetrated, refracted and reflected signals arrive at the receiving antenna, they superpose as a single signal sample, which is then extracted as Channel State Information~(CSI)~\cite{halperin2011tool}
\footnote{\url{https://en.wikipedia.org/wiki/Channel_state_information}}. As a result, the spatial information probed by WiFi signals is collapsed to a single CSI numeric, from which reconstructing the fine-grained spatial information of human body is an ill-posed problem. For instance, if we want to perceive human body in a $100\times 100$~px image coordinate (denoted by $\textbf{I}(P)$) from one CSI signal~(denoted by $H$), we have to solve $10^4$ unknowns given one $\textbf{I}(P)=f(H)$ equation.



We alleviate this problem by using the following two solutions: (1) Increasing the number of equations. In our person perception equipment, as shown in Figure~\ref{fig:wifi}~(b), we use $3$ transmitting antennas~($T$), $3$ receiving antennas~($R$) and $30$ EM frequencies ($F$). As a reward, the $3\times 3=9$ propagation pairs between antennas can capture the signals from different paths. The $30$ EM frequencies generate $30$ different superposing patterns at receiver antennas. This is because signals of different wavelengths can perceive objects at different scales. Moreover, we record $\textbf{I}$ as video frames at $20$~FPS and the CSI signals $\textbf{H}$ at $100$~Hz, such that each $\textbf{I}$ corresponds to $5$ sequential CSI samples. As a result, the system in Figure~\ref{fig:wifi}~(b) generates $3\times 3 \times 30 \times 5 = 1350$ equations of $\textbf{H}$ for one setting $\mathcal{W}(\cdot)$ of person~($P$) and environment~($E$). Our problem is reduced to learn a less ill-posed function $\textbf{I}(P)=f(\textbf{H})$, with $1350$ equations and $10^4$ unknowns. Note that the number of antennas, EM frequencies and CSI sampling rate are subject to IEEE~802.11n/ac WiFi communication standard and cannot be increased indefinitely. (2) Constraining the mapping complexity. We generate multiple spatial representations of person body from $\textbf{I}(P)$ and learn to map CSI to them using a multi-task DNNs. All these representations share the same spatial layout while highlight different body structures such as body mask, joints and limbs. This approach basically augments the data labels and further relieves the ill-posed problem.

\subsection{WiFi Signal, CSI and Hardware}

In the prevalent IEEE~802.11n/ac WiFi communication system, digital packages are carried in parallel by EM waves with multiple frequencies, called orthogonal frequency division multiplexing~(OFDM) technology. These packages are transmitted between multiple antenna pairs, called multiple-input-multiple-output~(MIMO). CSI is computed from signals between each pair of antennas at each frequency. A CSI sample, $c_i$, is computed as $ c_i = y_i/ x_i$, where $x_i$ and $y_i$ are the transmitted and received digital packages. Because of this, $c_i$ is irrelevant to the digital content of packages, but a measure of signal changes due to the reflection, refraction, absorption of EM wave with the person body and environment. Using CSI of WiFi, person perception is fundamentally possible.


To record CSI samples, we used a classic commercial WiFi Network Interface Card~(NIC)~\footnote{\url{http://a.co/d/bzh4tgb}} and leveraged an open source tool~\cite{halperin2011tool}, recorded CSI of $30$ EM waves with a bandwidth of $20$~MHz centering at the standard $2.4$~GHz WiFi. The $2.4$~GHz EM signal has a wavelength of around $12.5$~cm. Similar to standard house-hold WiFi routers, we uniformly spaced three receiver antennas within a wavelength, $12.5$~cm. This setting maximizes the difference of CSI captured at different receiver antennas. Figure~\ref{fig:csi_series} shows CSI samples corresponding to different person poses and locations under the same scene. 







%% file: tex/method.tex
\section{Deep Learning for Person-in-WiFi}

\subsection{Data and Annotations}

\begin{figure}[t]
\centering
\includegraphics[width=1.0\linewidth]{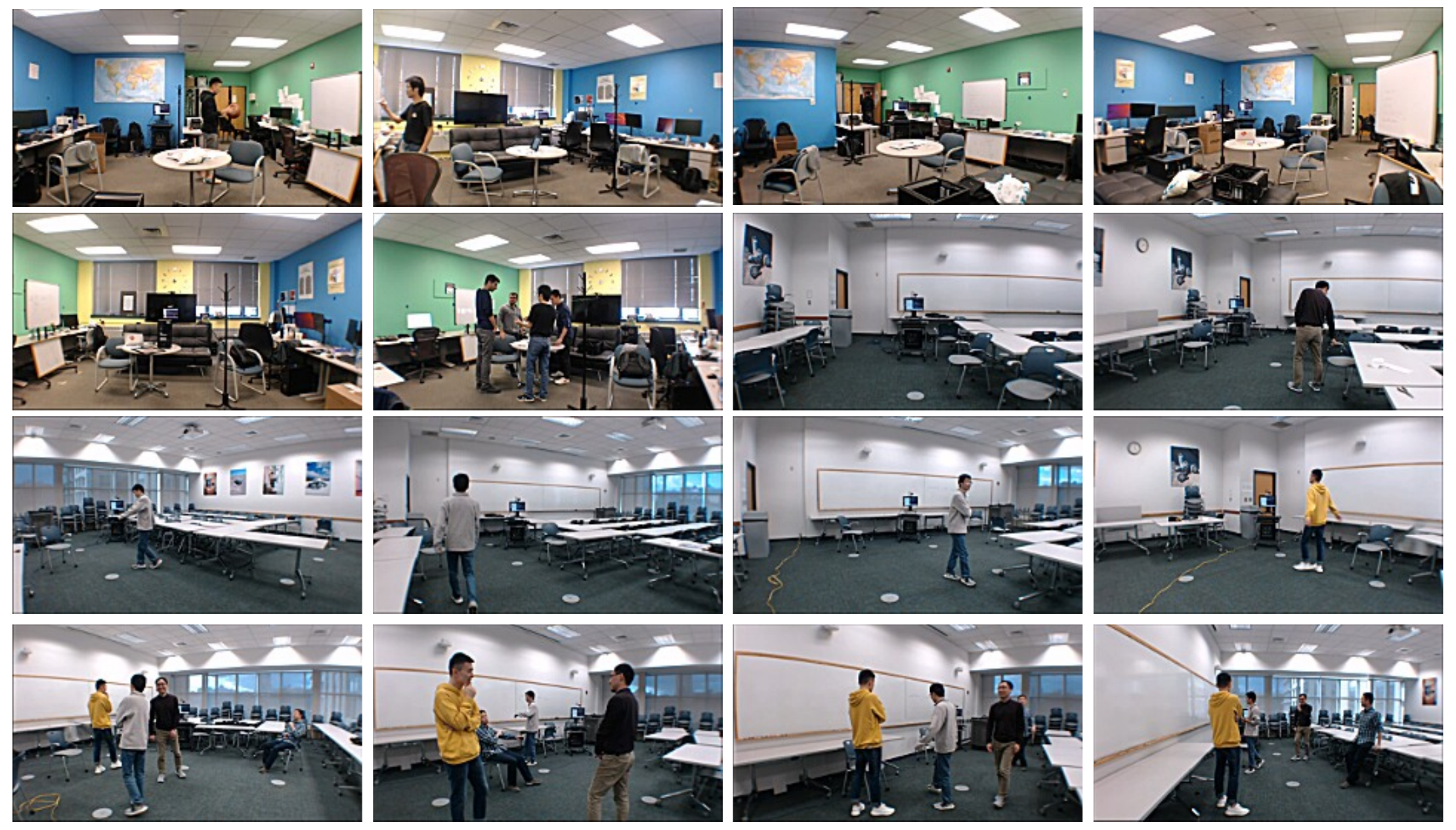}
\caption{Data collection under 16 indoor scenes and antenna locations.}\label{fig:scene}
\end{figure}

\begin{figure}[t]
\centering
\includegraphics[width=1.0\linewidth]{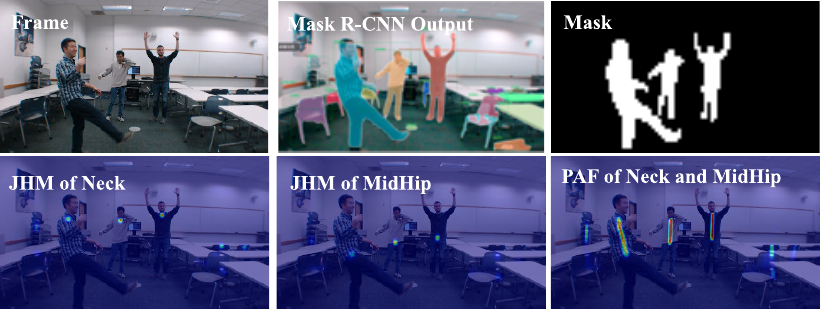}
\caption{Example of annotations from a video frame: body segmentation mask computed by Mask R-CNN~\cite{he2017mask}, JHMs and PAFs computed by OpenPose~\cite{cao2017realtime}.}\label{fig:annotation}
\end{figure}

\begin{figure*}[t]
\centering
\includegraphics[width=0.9\linewidth]{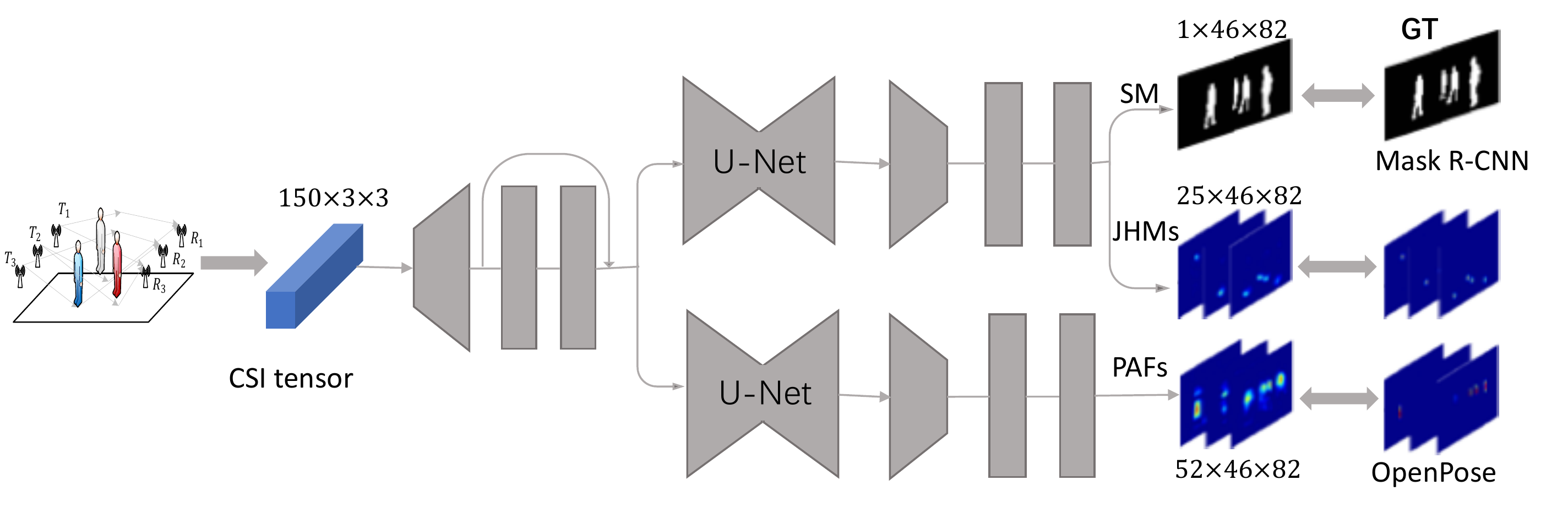}
\caption{Deep Neural Network for Person-in-WiFi: mapping from CSI to the body Segmentation Mask~(SM), Joint Heatmaps (JHMs) and Part Affinity Fields (PAFs).}\label{fig:network}
\end{figure*}

We recorded CSI at $100$~Hz from receiver antennas and videos at $20$~FPS from an RGB camera attached with receiver antennas. The videos are only used for annotating CSI. We synchronized CSI samples and video frames according to time stamps. In order to reduce the correlation between person body and environment, we collected data under 6 scenes in a laboratory office and 10 scenes in a classroom, shown in Figure~\ref{fig:scene}. Eight volunteers were asked to perform daily activities while the number of concurrent persons in the video varied from $1$ to $5$ (See Table~\ref{tab:data}). 
\begin{table}[t]
\small
\begin{tabular}{|  p{0.4cm}<{\centering}| p{0.8cm}<{\centering} | p{0.8cm}<{\centering} | p{0.8cm}<{\centering} | p{0.8cm}<{\centering} |p{0.75cm}<{\centering} |p{1.0cm}<{\centering}|}
\hline
$\#$P & 1     & 2     & 3     & 4     & 5    & Total    \\ \hline
$\#$F & 99,366 & 13,030 & 20,476 & 20,214 & 1,541 & 154,627 \\ \hline
\end{tabular}
\vspace{0.5pt}
\caption{Statistics of data: Number of concurrent persons~(\#P) and number of video frames~(\#F).}
\label{tab:data}
\end{table}

 From each video frame, we generated ground truth annotation for CSI as follows.
 For body segmentation, we used Mask R-CNN~\cite{he2017mask} to produce Segmentation Masks (SM) of persons, a $1\times 46 \times 82$ tensor, where $46$ and $82$ are height and width, respectively. For pose estimation, as explained in Section~\ref{sec:related}, we cannot use a person detector like Faster R-CNN~\cite{ren2015faster}, SSD~\cite{liu2016ssd} or Yolo~\cite{redmon2016you} to crop a person from the input CSI. We used the latest Body-25 model of OpenPose~\cite{cao2017realtime} to output body Joint Heat Maps (JHMs) and Part Affinity Fields~(PAFs). For each frame, JHMs is a $26\times 46 \times 82$ tensor, where the $26$ corresponds to $25$ joints and $1$ background. The PAFs is a $52\times 46 \times 82$ tensor where $52$ is for $x$ and $y$ coordinates of $26$ limbs. Figure~\ref{fig:annotation} shows examples of annotations.

\subsection{Networks}

Our deep neural networks (Figure~\ref{fig:network}) maps a CSI tensor to three output tensors: SM, JHMs and PAFs, where JHMs and PAFs are used later for the joint association as in~\cite{cao2017realtime}. 

The input tensor~($150\times 3 \times 3$) contains $5$ CSI samples corresponding to one video frame. The outputs are SM, JHMs and PAFs, all resized to $c\times 46 \times 82$. The input tensor is first upsampled to $150 \times 96 \times 96$, feed to a residual convolution block, and U-Nets~\cite{ronneberger2015u}\footnote{Any other alternatives of U-Nets can also be used in our network.}. U-Nets outputs are then downsampled to match ground truth using kernels with stride 2 on height and stride 1 on width. We found that SM (full body heatmaps) and JHMs (local joints/limbs heatmaps) are highly complementary, and one U-Net for SM and JHMs produced similar results as two independent U-Nets. 

We here go deeper and discuss how the spatial information embedded in CSI is reconstructed and mapped to SM, JHMs and PAFs. We interpret in the view of Receptive Field~(RF) of convolutional operation \cite{simonyan2014very}. Observe that dimensions of stacked CSI represent temporal information~($5$), EM frequency~($30$), and transmitting pairs among antennas~($3\times 3$), respectively. Because of the different relative distances and angles among transmitter and receiver antennas, the $3\times 3$ transmitting pairs capture 9 different 1D summaries of the same scene. Although the difference is subtle due to the small intervals comparing to distances to the human body, these 1D summaries are directly induced by the spatial layout of sensors. By reorganizing and reweighing, these $9$ numbers can potentially be to reconstruct 2D information of the scene. This is the reason we perform 2D convolution along the $3\times 3$ dimension of the input tensor. Observe that, the feature map after downsampling part of U-Nets has an RF size of $140$, which is larger than the height and width of the up-sampled $150 \times 96 \times 96$ tensor. This ensures that the feature maps in U-Nets observed all $9$ views among transmitter and receiver antennas. With supervision from annotations, the feature maps in U-Nets are forced to match the 2D spatial layout of the SM, JHMs and PAFs.  





\subsection{Loss and Matthew Weight}

\begin{figure}[t]
\centering
\includegraphics[width=1.0\linewidth]{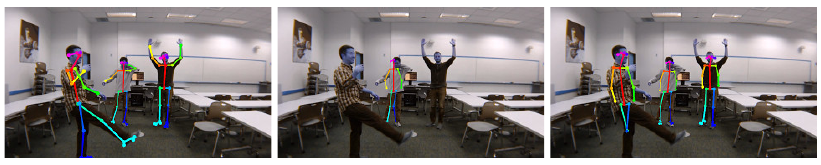}
\caption{Matthew Weight (MW) improves pose estimation. Left: ground-truth by OpenPose~\cite{cao2017realtime}; Middle: results with L2 loss; Right: results with L2 loss plus MW. }\label{fig:mattew}
\end{figure}

\begin{figure}[t]
\centering
\includegraphics[width=0.95\linewidth]{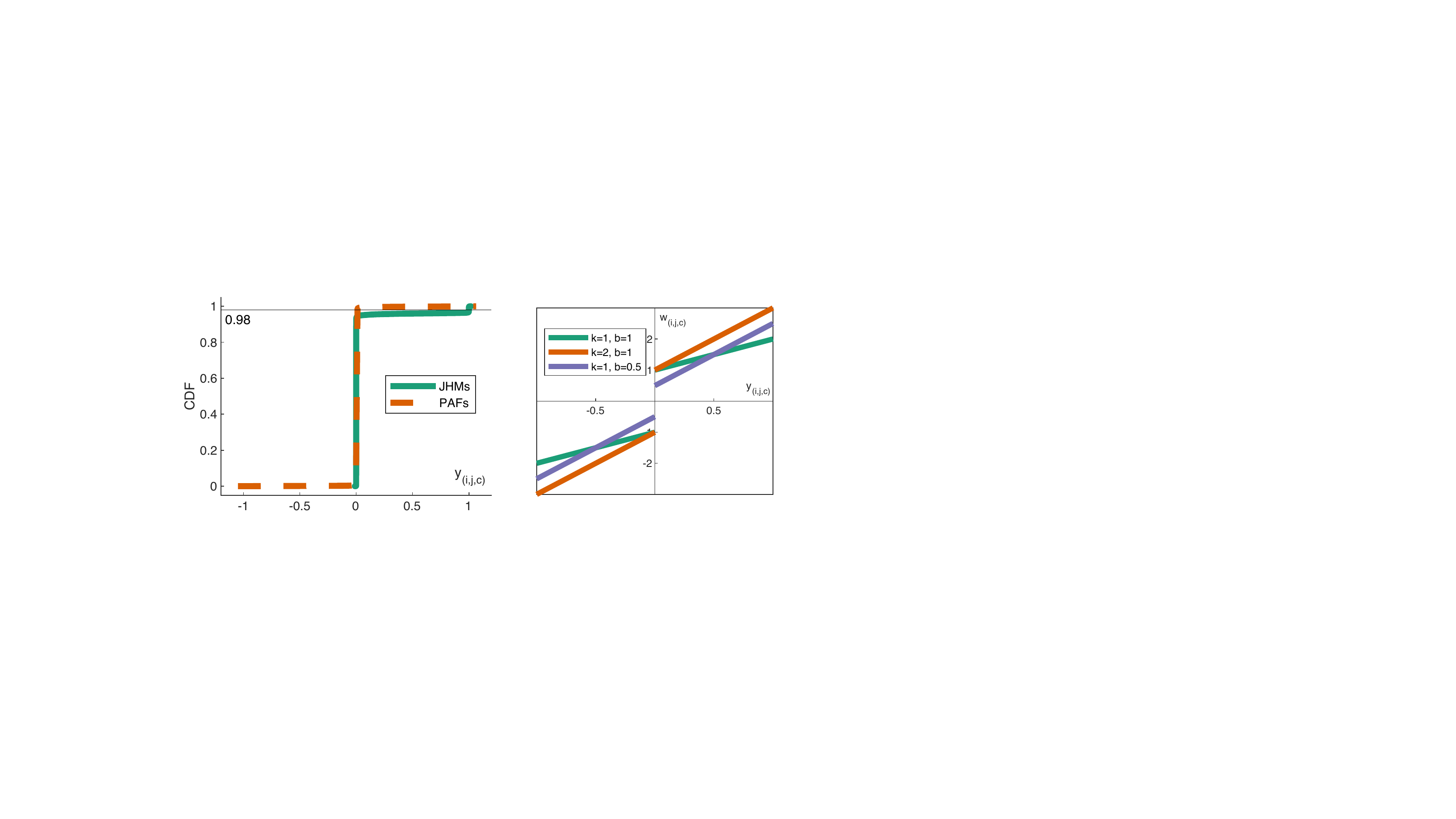}
\caption{Left: CDF of values of JHMs and PAFs. Right: examples of three Matthew Weight functions.}\label{fig:cdf}
\end{figure}

The network is trained over the sum of multiple losses
\begin{equation}
\mathcal{L} =  \lambda_1 L_\text{SM} + \lambda_2 L_\text{JHM} + \lambda_3 L_\text{PAF} 
\end{equation} 
where  $L_\text{SM}$  $L_\text{JHM}$ and $L_\text{PAF}$ are losses on body SM, JHMs and PAFs, respectively. $\lambda_i, i \in {1,2,3}$ are scalar weights to balance for these three losses. We use Binary Cross Entropy Loss to compute $L_\text{SM}$ as in \cite{he2017mask,ronneberger2015u,long2015fully}. Following \cite{cao2017realtime}, we set $\lambda_2$ and $\lambda_3$ as 1. $\lambda_1$ is empirically set as 0.1 to balance $L_\text{SM}$ with $L_\text{JHM}$ and $L_\text{PAF}$. Next, we go details about the problem we faced when optimizing $L_\text{JHM}$ and $L_\text{PAF}$, and the approach we proposed to tackle it.

Taking the JHMs loss as an example, directly using the popular L2 loss~\cite{chen2018cascaded,fang2017rmpe,papandreou2017towards,xiao2018simple} fails to generate good JHMs, see the middle of Figure~\ref{fig:mattew}. This is because the body joints only occupy very few pixels in the image, while L2 loss tends to average the regression error over all pixels. Figure~\ref{fig:cdf} shows the Cumulative Distribution Function~(CDF) of one JHMs tensor~($26\times 46 \times 82=98072$ scalars), showing that $98\%$ of the pixels are occupied by background, only less than 2\% are for joints. This problem could be partially relieved by multiple cascaded regression stages like OpenPose or Stacked Hourglass Networks~\cite{newell2016stacked}. Both solutions make networks much heavier. Leading top-down approaches focus on cropped person-wise features. But one cannot directly crop persons from CSI tensors. We use a simple but efficient loss to make networks pay more attention to body joints than the background:
\begin{equation}
L_\text{JHM}^{(i,j,c)} = w_{(i,j,c)} \cdot \left \|\hat{y}_{(i,j,c)} - y_{(i,j,c)}   \right \|_2^{2},
\end{equation}
where $w_{(i,j,c)}$ is the element-wise weight at index $(i,j,c)$, which is used to adjust optimizing attentions on JHMs;~$\hat{y}_{(i,j,c)}$ and $y_{(i,j,c)}$ are the prediction and annotation of JHMs at $(i,j,c)$. We propose to use the Matthew Weight (MW)~\footnote{We borrow the concept of Matthew Effect in Economics: the rich get richer, the poor get poorer.} to achieve the attention mechanism.
\begin{equation}
w_{(i,j,c)} =  k \cdot y_{(i,j,c)} + b \cdot \mathbbm{I}(y_{(i,j,c)}),
\label{eq:matthew}
\end{equation}
where $\mathbbm{I}(\cdot)$ outputs $+1$ when $y_{(i,j,c)} \geq 0$, otherwise $-1$. Figure~\ref{fig:cdf} are three MW examples. Note that MW is higher on larger elements (the body joints) in JHMs. Similarly, we applied MW in computing PAFs loss, $L_\text{PAF}$. Figure~\ref{fig:mattew} shows an example that MW significantly improves pose estimation comparing to directly using L2 loss.





\subsection{Implementation Details}
We implemented the networks in PyTorch. The batch size was 32, and the initial learning rate is 0.001. An Adam optimizer with $\beta_1=0.9$, $\beta_2=0.999$ was used in training. We used a $k=1, b=1$ MW in computing $L_\text{JHM}$ and a $k=1, b=0.3$ MW in computing $L_\text{PAF}$. The networks were trained for 20 epochs in all.

We used an OpenPose Python API~\footnote{\url{https://bit.ly/2zK3Aq5}} to conduct multi-person joint association given JHMs and PAFs. The output tensor is $p\times 25 \times 3$, where $p$ represents the number of persons that networks detected, $25\times 3$ denotes the $x$ axis, $y$ axis, and confidences of $25$ body joints. 






\begin{figure}[t]
\centering
\includegraphics[width=1.0\linewidth]{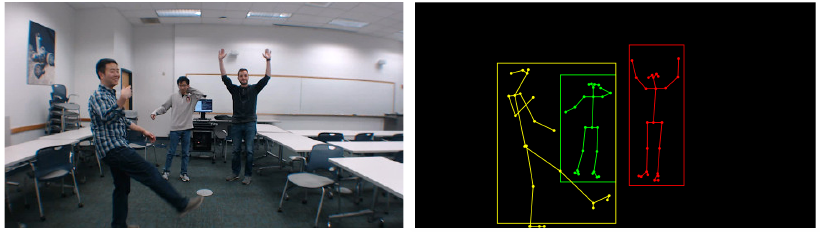}
\caption{Aligning body joints and person bounding-boxes for computing the PCK metric.}\label{fig:bbox}
\end{figure}

\begin{figure*}[t]
\centering
\includegraphics[width=1.0\linewidth]{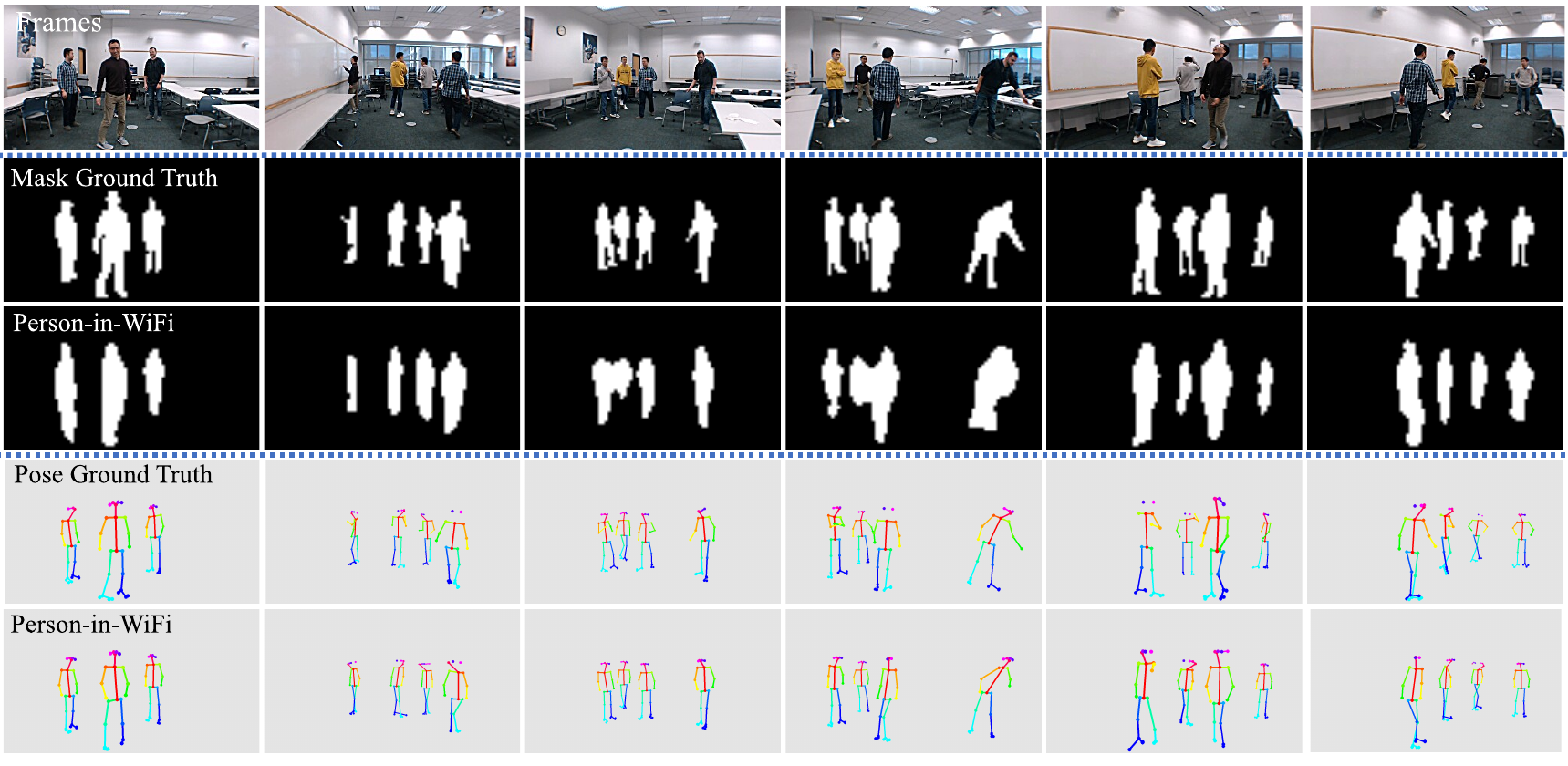}
\caption{Peson-in-WiFi results of body segmentation and pose estimation comparing to annotations by Mask R-CNN and and OpenPose.}\label{fig:all}
\end{figure*} 

\begin{table*}[t]
\small
\centering
\begin{tabular}{|c|c|c|c|c|c|c|c|c|c|c|c|}
\hline
\textbf{mIoU}&\textbf{mAP} & AP@50 & AP@55 & AP@60 & AP@65 & AP@70 & AP@75 & AP@80 & AP@85 & AP@90 & AP@95 \\ \hline
0.65& 0.38  & 0.91       & 0.85       & 0.75       & 0.59       & 0.40       & 0.20       & 0.07       & 0.01       & 0       & 0       \\ \hline
\end{tabular}
\vspace{5.0pt}
\caption{mIoU, mAP and APs of body segmentation. All metrics are the higher the better.}\label{tab:maskap}
\end{table*}

\begin{figure*}[t]
\centering
\includegraphics[width=0.24\linewidth]{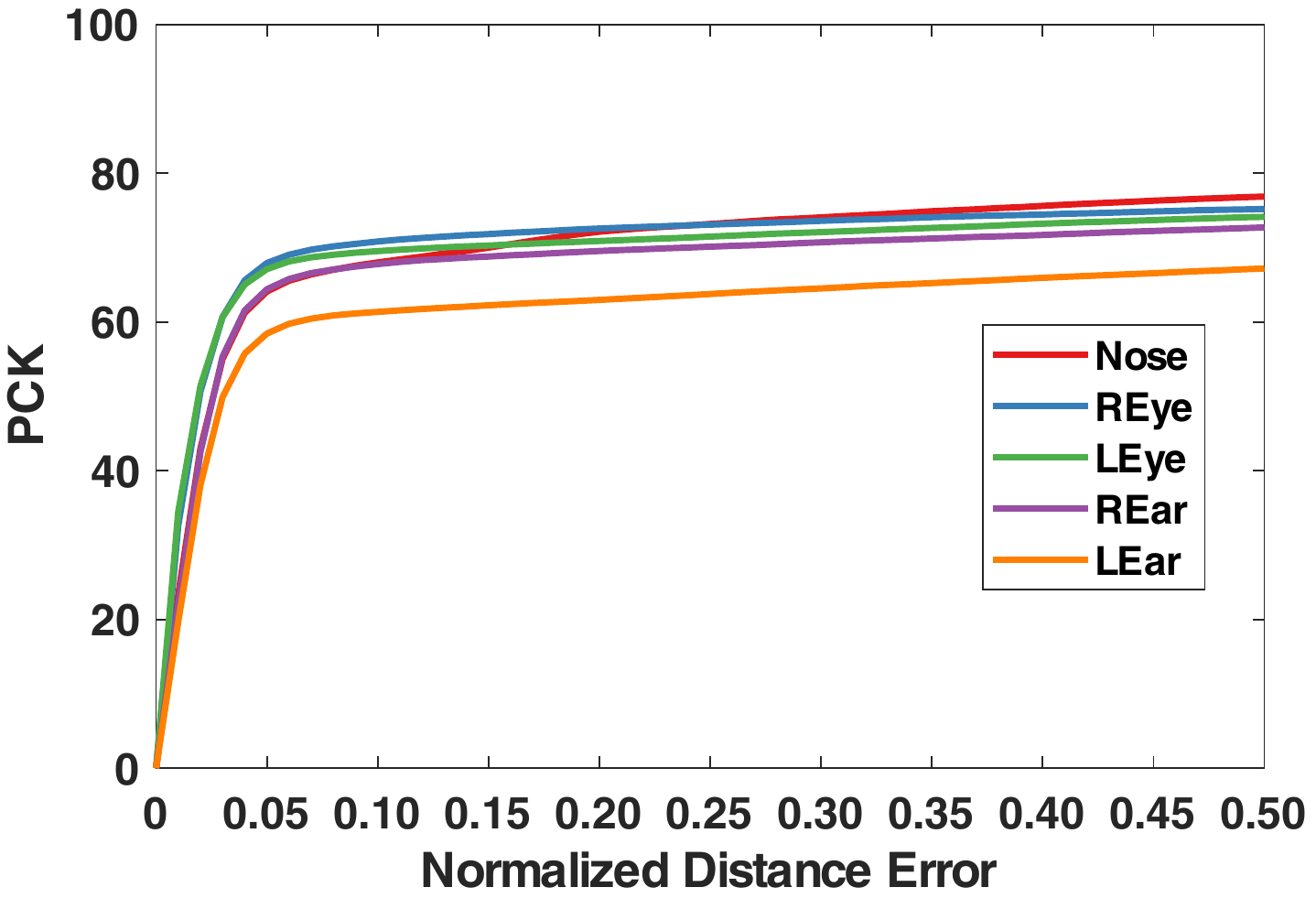}
\includegraphics[width=0.24\linewidth]{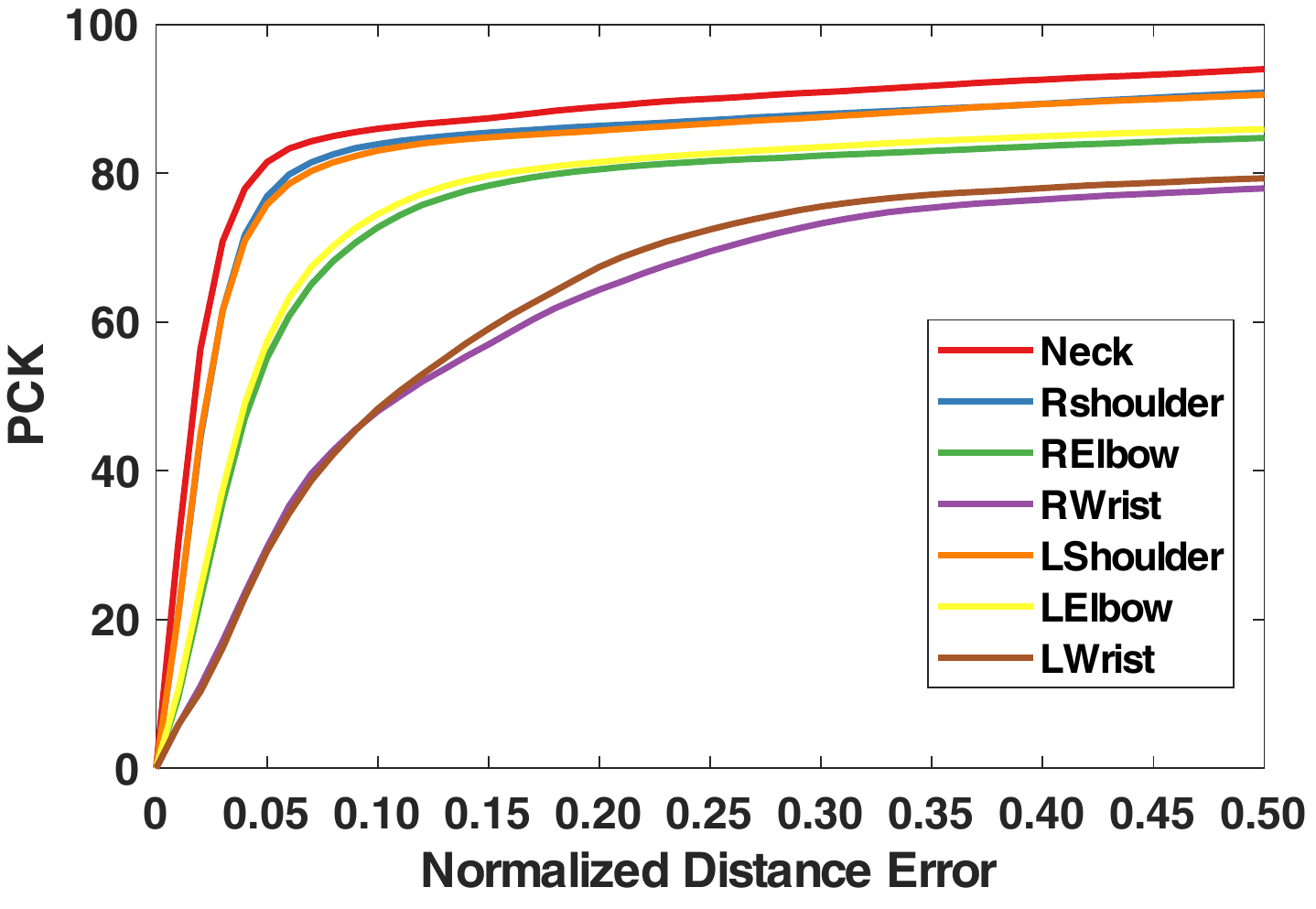}
\includegraphics[width=0.24\linewidth]{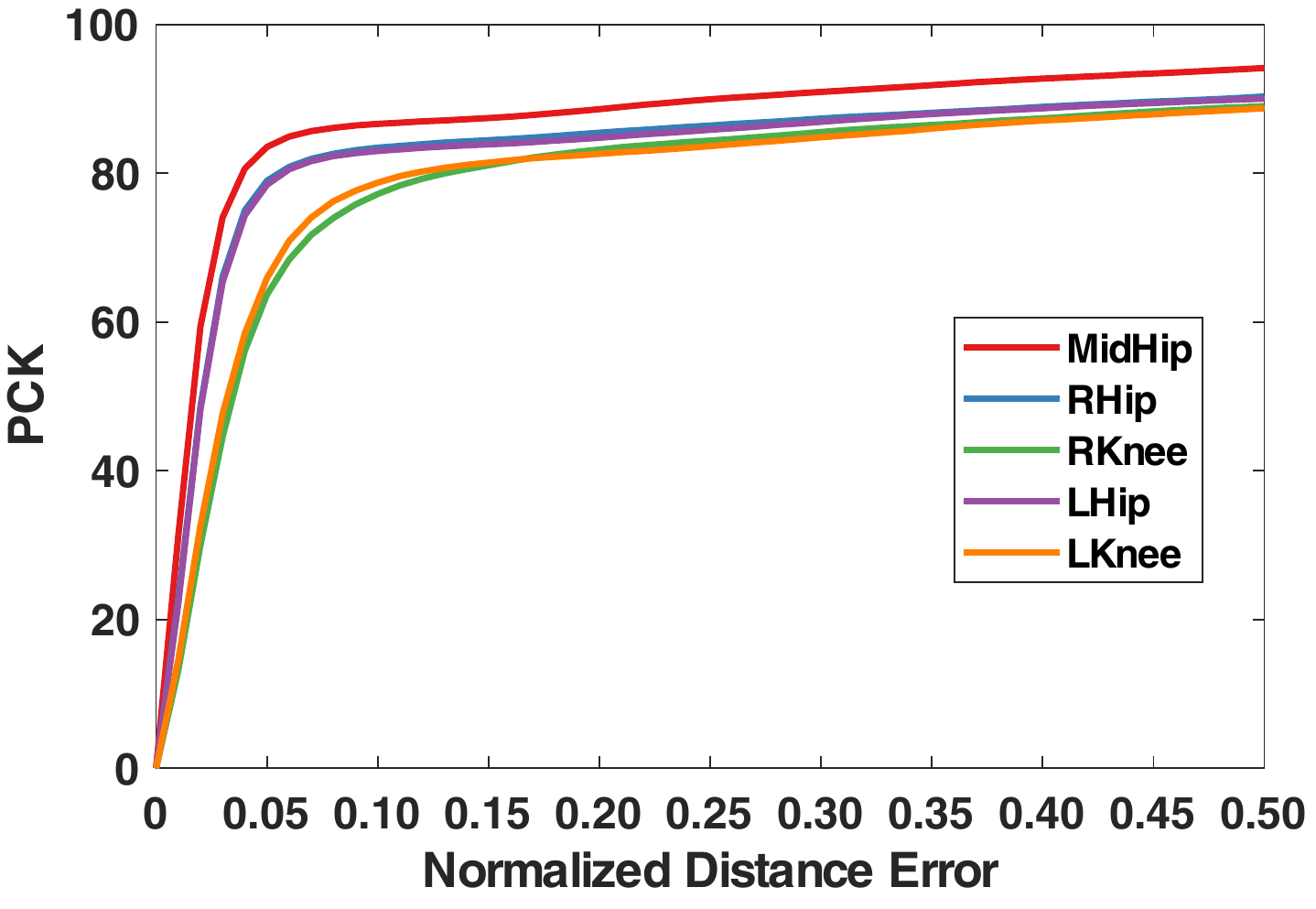}
\includegraphics[width=0.24\linewidth]{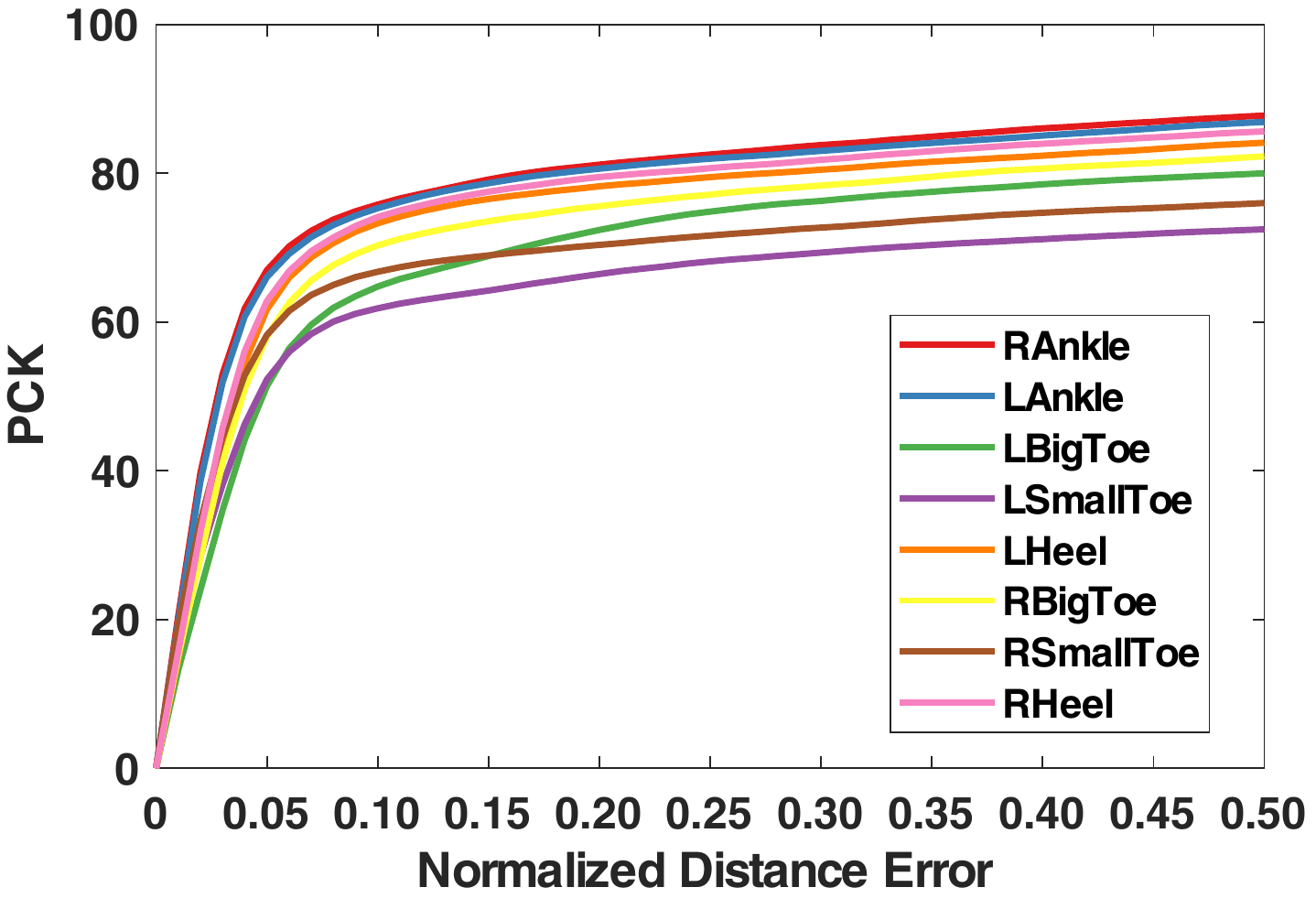}
\caption{PCKs of pose estimation. Horizontal axis: normalized distance error of joints (see Figure~\ref{fig:bbox} ). Vertical axis: PCKs of 25 body joints plot in four groups: (1) Head, (2) Torso\&Arms, (3) Legs, (4) Feet. PCKs are the higher the better.}\label{fig:posepck}
\end{figure*}

\begin{figure*}[t]
\centering
\includegraphics[width=1.0\linewidth]{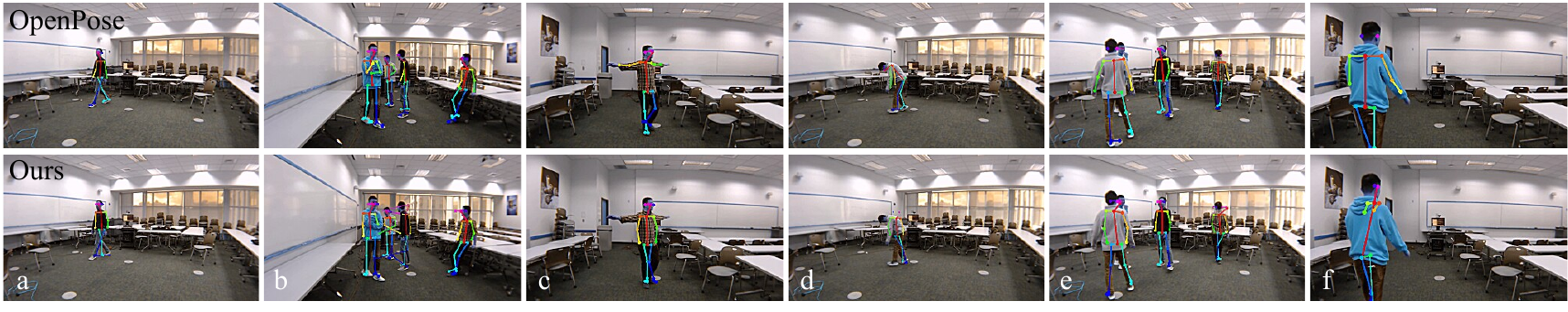}
\caption{Example of failure cases: (a-b) Lack of spatial resolution; (c-d) Rare poses; (e-f) Incomplete annotations from camera view. }\label{fig:failures}
\vspace{-0.3cm}
\end{figure*}

%% file: tex/evaluation.tex
\section{Experiments}


Data were collected by groups of subjects~(1-5 persons per group). Each group was asked to perform a continuous motion in the scene. We used automatic annotations for segmentation~(Mask~R-CNN) and poses~(OpenPose) on mono-camera images that were synchronized with CSI samples. Note that this is a  proof-of-concept experiments and can be further improved by high-quality manual annotations and multi-camera images~(for occlusions or behind a wall).

\textbf{Body Segmentation Metrics:} Mean Intersection over Union (mIoU) and mAP (over AP@50 to AP@95) as used in the COCO challenge, where: 
\begin{equation}
    \mathrm{AP@}a = \frac{1}{N} \sum_{n=1}^{N}\mathbbm{I}(100 \cdot IOU_n \geq a)
    \label{eq:ap}
\end{equation}
where $N$ is the number of test frames, and $\mathbbm{I}$ is a logical operation which outputs 1 if True and outputs 0 if False. All metrics are the higher the better.

\textbf{Pose Estimation Metrics: } Percentage of Correct Keypoint~(PCK)~\cite{andriluka20142d,yang2013articulated,newell2016stacked}. We made a slight modification as~Equation ~\ref{eq:pck} considering annotations we have. 
\begin{equation}
    \mathrm{PCK_i@}a =  \frac{1}{P} \sum_{p=1}^{P}\mathbbm{I}\left(\frac{\left \|pd_i^p-gt_i^p  \right \|^2_{2} }{\sqrt[2]{w^{p2}+h^{p2}}} \leq a \right),
    \label{eq:pck}
\end{equation}
where $\mathbbm{I}$ are the same as Equation~\ref{eq:ap}. $P$ is the amount of persons in test frames. $i$ denotes the index of body joint and $i \in \left \{1,2,...,25 \right \}$. $\left \|pd_i^p-gt_i^p  \right \|^2_{2}$ is the Euclidean pixel distance between the prediction and ground-truth, which is normalized by the diagonal length of the person bounding box, $\sqrt[2]{w^{p2}+h^{p2}}$. To get person bounding boxes, we aligned body joint coordinates from OpenPose~\cite{cao2017realtime} with the bounding box from Mask R-CNN~\cite{he2017mask} (see Figure~\ref{fig:bbox}). 

We did not use the Object Keypoint Similarity~(OKS)~AP@$a$ of the COCO Keypoint Detection challenge for two reasons: (1) Our 25 body joints requires $25$ hyper-parameters to compute OKS, but the COCO dataset only provides $18$; (2) The COCO dataset hyper-parameters are based on statistics of COCO data and may introduce bias in evaluating our dataset.

In the first experiment, the first $80\%$ of samples of each subject group were used for training and the later $20\%$ for testing. The training and testing samples are different in locomotion and body poses, but share the person identities and environments. The amount of training/test samples are $123631$ and $30996$, respectively.

\subsection{Performance of Body Segmentation}
The mAP over AP@50-AP@95 of body segmentation is $0.38$ (see Table~\ref{tab:maskap}). High values of AP@50-AP@70 mean that person profiles can be properly detected from WiFi signals. Low values of AP@80-AP@95 indicate that subtle body masks are not well-detected. Figure~\ref{fig:all} qualitatively show masks from WiFi comparing to the annotations by Mask R-CNN~\cite{he2017mask}. Most body locations, torsos, legs can be well-segmented, which is good enough for safety applications such as detecting falling of elderly \cite{wang2017wifall} and physical conflicts among people. 

\subsection{Performance of Pose Estimation}
Since we used Body-25 model of OpenPose to annotate the poses,  $25$ PCKs were computed for the 25 body joints. We plot PCKs in 4 groups in Figure~\ref{fig:posepck} and analyze the performance of pose estimation. The 4 groups of joints are Head~\{Nose, REye, LEye, REar, LEar\}, Torso\&Arms~\{Neck, Rshoulder, RElbow, RWrist, LShoulder, LElbow, LWrist\}, Legs~\{MidHip, RHip, RKnee, LHip, LKnee\} and Feet~\{RAnkle, LAnkle, LBigToe, LSmallToe, LHeel, RBigToe, RSmallToe, RHeel\}. 

As shown in Figure~\ref{fig:posepck}, the estimation of most joints produced high PCKs (vertical axis) at low ($0.1$) normalized distance error (horizontal axis). In other word, most joints were located within less than $0.1$ of diagonal length of the person bounding box. Generally, joints of large body parts like in group Torso\&Arms and group Legs have higher PCKs, while joints in group Head or group Feet tend to have lower PCKs. We will analyze the failure cases in the next subsection. Figure~\ref{fig:all} show pose estimation achieved using WiFi comparing to annotations from OpenPose.

\subsection{Gaps with Camera-based Approaches}

\begin{table}[h]
\footnotesize\centering
\begin{tabular}{|c|c|c|}\hline
             & mIoU & mPCK@0.20\\ 
\hline Person-in-WiFi & 0.66 & 78.75\\
\hline
Mask-RCNN &  0.83   & -\\
\hline
OpenPose  &  - &  89.48\\
\hline
\end{tabular}
\caption{Gaps between Person-in-WiFi (Trained on annotations of camera-based approaches) and camera-based approaches.}
\label{tab:compare_with_camera}
\vspace{-0.2cm}
\end{table}
Above Person-in-WiFi models were trained on, therefore bounded by the annotations produced using Mask R-CNN and OpenPose. It is still possible to evaluate the gaps between two perception approaches. Table.~\ref{tab:compare_with_camera} compares the results on $160$ samples that were uniformly selected from above test set and manually annotated~\footnote{Labeling tool:~\url{https://github.com/wkentaro/labelme}}. The quantitative gaps are noticeable, but could be reduced with more data and high-quality annotations, considering that Mask R-CNN and OpenPose were trained with abundant data. 

\subsection{Failure cases}
Several failure cases exist in our current results (see Figure~\ref{fig:failures}) 
(1) Lack of spatial resolution (See Figure~\ref{fig:failures} (a-b)). Small limbs may be bypassed or mixed in WiFi EM waves due to the diffraction effect. For instance, WiFi signals at $2.4$~GHz have a wavelength of around $12.5$~cm, and may miss an object of less than $12.5$~cm along its direct propagation path. However, multiple propagation paths by $3$ receiver antennas and countless reflection paths of signals can capture the trace of small limbs. Figure~\ref{fig:all} showed many successful cases. The failures could be improved by higher weights on regression errors of small limbs, more data and temporal smoothing. (2) Rare poses (Figure~\ref{fig:failures} (c-d)). More data and random data augmentation can improve the results. (3) Incomplete annotations: Camera has narrower field-of-view ($70^\circ$ horizontally) than the WiFi antennas that broadcast signals in $360^\circ$. Annotations from a single camera is incomplete on occluded body parts (Figure~\ref{fig:failures} (e-f)). Annotation with multi-camera videos could address the issues.

\begin{figure}[H]
\centering
\includegraphics[width=1.0\linewidth]{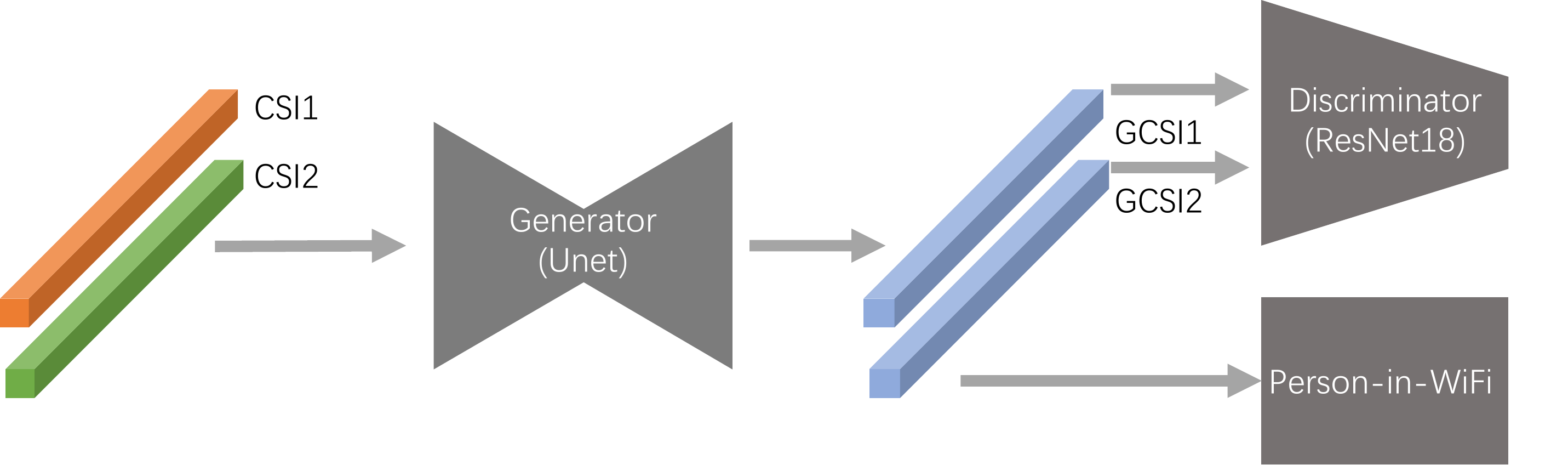}
\caption{Adversarial training for environment invariance.}\label{fig:gan}
\vspace{-0.3cm}
\end{figure}

\subsection{Deployment in Untrained Environment}
WiFi signals in different environments exhibit significantly different prorogation patterns. It is still an open problem, yet possible to deploy WiFi-based perception system in an untrained environment. The only work we found to address this issue was for activity classification~\cite{jiang2018towards}, a much simpler task than Person-in-WiFi. 

As a preliminary attempt to deploy Person-in-WiFi to untrained environment, we developed a GAN-based training approach:
\textbf{Step 1:} pre-training a binary environment discriminator (D) which takes a random pair of CSI tensors as inputs, and produces $1$ if the paired tensors are from a same environment, and $0$ otherwise; \textbf{Step 2:} training the network in Fig.~\ref{fig:gan}; Fixing discriminator~(D) in Step 1; updating a Unet generator network~(G), such that any pairs of generator outputs~(GCSI) produce $1$s~(same environment). Meanwhile, GCSI tensors are used as input tensors of the Person-in-WiFi network~(see~Fig.~\ref{fig:network}). The generator and Person-in-WiFi network are updated simultaneously.

We conducted preliminary experiments on 14 training scenes and 2 testing scenes. Above training approach improved segmentation mIoU from $0.12$ to $0.24$, improved pose estimation mPCK@0.20 from $19.34$ to $31.06$. Nevertheless, further improvement on untrained environment requires more data and annotations. To encourage researchers to explore WiFi-based fine-grained person perception, we plan to release our current dataset for research purposes and will continue to build a large scale dataset.



%% file: egpaper_final.bbl
\begin{thebibliography}{10}\itemsep=-1pt

\bibitem{adib2015capturing}
F.~Adib, C.-Y. Hsu, H.~Mao, D.~Katabi, and F.~Durand.
\newblock Capturing the human figure through a wall.
\newblock {\em TOG}, 34(6):219, 2015.

\bibitem{adib20143d}
F.~Adib, Z.~Kabelac, D.~Katabi, and R.~C. Miller.
\newblock 3d tracking via body radio reflections.
\newblock In {\em NSDI}, volume~14, pages 317--329, 2014.

\bibitem{adib2013see}
F.~Adib and D.~Katabi.
\newblock See through walls with wifi!
\newblock {\em SIGCOMM Comput. Commun. Rev.}, 43(4):75--86, Aug. 2013.

\bibitem{ali2015keystroke}
K.~Ali, A.~X. Liu, W.~Wang, and M.~Shahzad.
\newblock Keystroke recognition using wifi signals.
\newblock In {\em MobiCom}, pages 90--102. ACM, 2015.

\bibitem{andriluka20142d}
M.~Andriluka, L.~Pishchulin, P.~Gehler, and B.~Schiele.
\newblock 2d human pose estimation: New benchmark and state of the art
  analysis.
\newblock In {\em CVPR}, pages 3686--3693, 2014.

\bibitem{benedek20143d}
C.~Benedek.
\newblock 3d people surveillance on range data sequences of a rotating lidar.
\newblock {\em Pattern Recognition Letters}, 50:149--158, 2014.

\bibitem{benedek2018lidar}
C.~Benedek, B.~G{\'a}lai, B.~Nagy, and Z.~Jank{\'o}.
\newblock Lidar-based gait analysis and activity recognition in a 4d
  surveillance system.
\newblock {\em TCSVT}, 28(1):101--113, 2018.

\bibitem{cao2017realtime}
Z.~Cao, T.~Simon, S.-E. Wei, and Y.~Sheikh.
\newblock Realtime multi-person 2d pose estimation using part affinity fields.
\newblock In {\em CVPR}, 2017.

\bibitem{chen2014articulated}
X.~Chen and A.~L. Yuille.
\newblock Articulated pose estimation by a graphical model with image dependent
  pairwise relations.
\newblock In {\em NIPS}, pages 1736--1744, 2014.

\bibitem{chen2018cascaded}
Y.~Chen, Z.~Wang, Y.~Peng, Z.~Zhang, G.~Yu, and J.~Sun.
\newblock Cascaded pyramid network for multi-person pose estimation.
\newblock {\em arXiv preprint}, 2018.

\bibitem{costea2017fast}
A.~D. Costea, R.~Varga, and S.~Nedevschi.
\newblock Fast boosting based detection using scale invariant multimodal
  multiresolution filtered features.
\newblock In {\em CVPR}, pages 993--1002, July 2017.

\bibitem{droeschel2018efficientCS}
D.~Droeschel and S.~Behnke.
\newblock Efficient continuous-time slam for 3d lidar-based online mapping.
\newblock {\em ICRA}, pages 1--9, 2018.

\bibitem{fan2015combining}
X.~Fan, K.~Zheng, Y.~Lin, and S.~Wang.
\newblock Combining local appearance and holistic view: Dual-source deep neural
  networks for human pose estimation.
\newblock In {\em CVPR}, pages 1347--1355, 2015.

\bibitem{fang2017rmpe}
H.~Fang, S.~Xie, Y.-W. Tai, and C.~Lu.
\newblock Rmpe: Regional multi-person pose estimation.
\newblock In {\em ICCV}, volume~2, 2017.

\bibitem{fragkiadaki2015recurrent}
K.~Fragkiadaki, S.~Levine, P.~Felsen, and J.~Malik.
\newblock Recurrent network models for human dynamics.
\newblock In {\em ICCV}, pages 4346--4354, 2015.

\bibitem{halperin2011tool}
D.~Halperin, W.~Hu, A.~Sheth, and D.~Wetherall.
\newblock Tool release: Gathering 802.11 n traces with channel state
  information.
\newblock {\em ACM SIGCOMM Computer Communication Review}, 41(1):53--53, 2011.

\bibitem{han2015realtime}
X.~Han, J.~Lu, Y.~Tai, and C.~Zhao.
\newblock A real-time lidar and vision based pedestrian detection system for
  unmanned ground vehicles.
\newblock In {\em ACPR}, pages 635--639, Nov 2015.

\bibitem{he2017mask}
K.~He, G.~Gkioxari, P.~Doll{\'a}r, and R.~Girshick.
\newblock Mask r-cnn.
\newblock In {\em ICCV}, pages 2980--2988. IEEE, 2017.

\bibitem{hess2016real}
W.~Hess, D.~Kohler, H.~Rapp, and D.~Andor.
\newblock Real-time loop closure in 2d lidar slam.
\newblock In {\em ICRA}, pages 1271--1278, May 2016.

\bibitem{holl2017holography}
P.~M. Holl and F.~Reinhard.
\newblock Holography of wi-fi radiation.
\newblock {\em Physical review letters}, 118(18):183901, 2017.

\bibitem{huang2014feasibility}
D.~Huang, R.~Nandakumar, and S.~Gollakota.
\newblock Feasibility and limits of wi-fi imaging.
\newblock In {\em SenSys}, pages 266--279. ACM, 2014.

\bibitem{jain2014modeep}
A.~Jain, J.~Tompson, Y.~LeCun, and C.~Bregler.
\newblock Modeep: A deep learning framework using motion features for human
  pose estimation.
\newblock In {\em ACCV}, pages 302--315. Springer, 2014.

\bibitem{jiang2018towards}
W.~Jiang, C.~Miao, F.~Ma, S.~Yao, Y.~Wang, Y.~Yuan, H.~Xue, C.~Song, X.~Ma,
  D.~Koutsonikolas, et~al.
\newblock Towards environment independent device free human activity
  recognition.
\newblock In {\em MobiCom}, pages 289--304. ACM, 2018.

\bibitem{kotaru2015spotfi}
M.~Kotaru, K.~Joshi, D.~Bharadia, and S.~Katti.
\newblock Spotfi: Decimeter level localization using wifi.
\newblock {\em SIGCOMM Comput. Commun. Rev.}, 45(4):269--282, Aug. 2015.

\bibitem{leigh2015person}
A.~Leigh, J.~Pineau, N.~Olmedo, and H.~Zhang.
\newblock Person tracking and following with 2d laser scanners.
\newblock In {\em ICRA}, pages 726--733. IEEE, 2015.

\bibitem{li2016csi}
M.~Li, Y.~Meng, J.~Liu, H.~Zhu, X.~Liang, Y.~Liu, and N.~Ruan.
\newblock When csi meets public wifi: Inferring your mobile phone password via
  wifi signals.
\newblock In {\em CCS}, pages 1068--1079. ACM, 2016.

\bibitem{lin2017feature}
T.~Lin, P.~Dollár, R.~Girshick, K.~He, B.~Hariharan, and S.~Belongie.
\newblock Feature pyramid networks for object detection.
\newblock In {\em CVPR}, pages 936--944, July 2017.

\bibitem{liu2016ssd}
W.~Liu, D.~Anguelov, D.~Erhan, C.~Szegedy, S.~Reed, C.-Y. Fu, and A.~C. Berg.
\newblock Ssd: Single shot multibox detector.
\newblock In {\em ECCV}, pages 21--37. Springer, 2016.

\bibitem{long2015fully}
J.~Long, E.~Shelhamer, and T.~Darrell.
\newblock Fully convolutional networks for semantic segmentation.
\newblock In {\em CVPR}, pages 3431--3440, 2015.

\bibitem{matti2017combining}
D.~Matti, H.~K. Ekenel, and J.~Thiran.
\newblock Combining lidar space clustering and convolutional neural networks
  for pedestrian detection.
\newblock {\em CoRR}, abs/1710.06160, 2017.

\bibitem{maturana2015voxnet}
D.~Maturana and S.~Scherer.
\newblock Voxnet: A 3d convolutional neural network for real-time object
  recognition.
\newblock In {\em IROS}, pages 922--928. IEEE, 2015.

\bibitem{newell2016stacked}
A.~Newell, K.~Yang, and J.~Deng.
\newblock Stacked hourglass networks for human pose estimation.
\newblock In {\em ECCV}, pages 483--499. Springer, 2016.

\bibitem{papandreou2017towards}
G.~Papandreou, T.~Zhu, N.~Kanazawa, A.~Toshev, J.~Tompson, C.~Bregler, and
  K.~Murphy.
\newblock Towards accurate multi-person pose estimation in the wild.
\newblock In {\em CVPR}, pages 3711--3719, July 2017.

\bibitem{pfister2015flowing}
T.~Pfister, J.~Charles, and A.~Zisserman.
\newblock Flowing convnets for human pose estimation in videos.
\newblock In {\em ICCV}, pages 1913--1921, 2015.

\bibitem{pishchulin2016deepcut}
L.~Pishchulin, E.~Insafutdinov, S.~Tang, B.~Andres, M.~Andriluka, P.~V. Gehler,
  and B.~Schiele.
\newblock Deepcut: Joint subset partition and labeling for multi person pose
  estimation.
\newblock In {\em CVPR}, pages 4929--4937, 2016.

\bibitem{premebida2014pedestrian}
C.~Premebida, J.~Carreira, J.~Batista, and U.~Nunes.
\newblock Pedestrian detection combining rgb and dense lidar data.
\newblock In {\em IROS}, pages 4112--4117, Sept 2014.

\bibitem{pu2013whole}
Q.~Pu, S.~Gupta, S.~Gollakota, and S.~Patel.
\newblock Whole-home gesture recognition using wireless signals.
\newblock In {\em MobiCom}, pages 27--38. ACM, 2013.

\bibitem{qian2017inferring}
K.~Qian, C.~Wu, Z.~Zhou, Y.~Zheng, Z.~Yang, and Y.~Liu.
\newblock Inferring motion direction using commodity wi-fi for interactive
  exergames.
\newblock In {\em CHI}, pages 1961--1972. ACM, 2017.

\bibitem{redmon2016you}
J.~Redmon, S.~Divvala, R.~Girshick, and A.~Farhadi.
\newblock You only look once: Unified, real-time object detection.
\newblock In {\em CVPR}, pages 779--788, 2016.

\bibitem{ren2015faster}
S.~Ren, K.~He, R.~Girshick, and J.~Sun.
\newblock Faster r-cnn: Towards real-time object detection with region proposal
  networks.
\newblock In {\em NIPS}, pages 91--99, 2015.

\bibitem{ronneberger2015u}
O.~Ronneberger, P.~Fischer, and T.~Brox.
\newblock U-net: Convolutional networks for biomedical image segmentation.
\newblock In {\em International Conference on Medical image computing and
  computer-assisted intervention}, pages 234--241. Springer, 2015.

\bibitem{shackleton2010tracking}
J.~Shackleton, B.~VanVoorst, and J.~Hesch.
\newblock Tracking people with a 360-degree lidar.
\newblock In {\em AVSS}, pages 420--426, Aug 2010.

\bibitem{shaker2008fuzzy}
S.~Shaker, J.~J. Saade, and D.~Asmar.
\newblock Fuzzy inference-based person-following robot.
\newblock {\em International Journal of Systems Applications, Engineering and
  Development}, 2(1):29--34, 2008.

\bibitem{simonyan2014very}
K.~Simonyan and A.~Zisserman.
\newblock Very deep convolutional networks for large-scale image recognition.
\newblock {\em arXiv preprint arXiv:1409.1556}, 2014.

\bibitem{tompson2014joint}
J.~J. Tompson, A.~Jain, Y.~LeCun, and C.~Bregler.
\newblock Joint training of a convolutional network and a graphical model for
  human pose estimation.
\newblock In {\em NIPS}, pages 1799--1807, 2014.

\bibitem{toshev2014deeppose}
A.~Toshev and C.~Szegedy.
\newblock Deeppose: Human pose estimation via deep neural networks.
\newblock In {\em CVPR}, pages 1653--1660, 2014.

\bibitem{wang2018csi}
F.~Wang, J.~Han, S.~Zhang, X.~He, and D.~Huang.
\newblock Csi-net: Unified human body characterization and action recognition.
\newblock {\em arXiv preprint arXiv:1810.03064}, 2018.

\bibitem{wang2016human}
H.~Wang, D.~Zhang, J.~Ma, Y.~Wang, Y.~Wang, D.~Wu, T.~Gu, and B.~Xie.
\newblock Human respiration detection with commodity wifi devices: do user
  location and body orientation matter?
\newblock In {\em Ubicomp}, pages 25--36. ACM, 2016.

\bibitem{wang2017wifall}
Y.~Wang, K.~Wu, and L.~M. Ni.
\newblock Wifall: Device-free fall detection by wireless networks.
\newblock {\em TMC}, 16(2):581--594, 2017.

\bibitem{wang2017characterization}
Z.~Wang, Y.~Liu, Q.~Liao, H.~Ye, M.~Liu, and L.~Wang.
\newblock Characterization of a rs-lidar for 3d perception.
\newblock {\em arXiv preprint arXiv:1709.07641}, 2017.

\bibitem{wei2016convolutional}
S.-E. Wei, V.~Ramakrishna, T.~Kanade, and Y.~Sheikh.
\newblock Convolutional pose machines.
\newblock In {\em ICCV}, pages 4724--4732, 2016.

\bibitem{xiao2018simple}
B.~Xiao, H.~Wu, and Y.~Wei.
\newblock Simple baselines for human pose estimation and tracking.
\newblock {\em arXiv preprint arXiv:1804.06208}, 2018.

\bibitem{yang2015wifi}
C.~Yang and H.-R. Shao.
\newblock Wifi-based indoor positioning.
\newblock {\em IEEE Communications Magazine}, 53(3):150--157, 2015.

\bibitem{yang2013articulated}
Y.~Yang and D.~Ramanan.
\newblock Articulated human detection with flexible mixtures of parts.
\newblock {\em TPAMI}, 35(12):2878--2890, 2013.

\bibitem{zhao2018through}
M.~Zhao, T.~Li, M.~Abu~Alsheikh, Y.~Tian, H.~Zhao, A.~Torralba, and D.~Katabi.
\newblock Through-wall human pose estimation using radio signals.
\newblock In {\em CVPR}, pages 7356--7365, 2018.

\bibitem{zhou2016sparseness}
X.~Zhou, M.~Zhu, S.~Leonardos, K.~G. Derpanis, and K.~Daniilidis.
\newblock Sparseness meets deepness: 3d human pose estimation from monocular
  video.
\newblock In {\em CVPR}, pages 4966--4975, 2016.

\end{thebibliography}
